# Deep Learning-based Compressed Domain Multimedia for Man and Machine: A Taxonomy and Application to Point Cloud Classification

Abdelrahman Seleem[1, 2, 4], Student Member, IEEE, André F. R. Guarda[2], Member, IEEE, Nuno M. M. Rodrigues[2, 3], Senior Member, IEEE, Fernando Pereira[1, 2], Fellow, IEEE

[1]Instituto Superior Técnico - Universidade de Lisboa, Lisbon, Portugal
[2]Instituto de Telecomunicações, Portugal
[3]ESTG, Politécnico de Leiria, Leiria, Portugal
[4]Faculty of Computers and Information, South Valley University, Qena, Egypt

Corresponding author: Abdelrahman Seleem (e-mail: a.seleem@lx.it.pt).

This work was funded by the Fundação para a Ciência e a Tecnologia (FCT, Portugal) through the research project PTDC/EEI-COM/1125/2021, entitled "Deep Learning-based Point Cloud Representation".

**ABSTRACT** In the current golden age of multimedia, human visualization is no longer the single main target, with the final consumer often being a machine which performs some processing or computer vision tasks. In both cases, deep learning plays a fundamental role in extracting features from the multimedia representation data, usually producing a compressed representation referred to as latent representation. The increasing development and adoption of deep learning-based solutions in a wide area of multimedia applications have opened an exciting new vision where a common compressed multimedia representation is used for both man and machine. The main benefits of this vision are two-fold: i) improved performance for the computer vision tasks, since the effects of coding artifacts are mitigated; and ii) reduced computational complexity, since prior decoding is not required. This paper proposes the first taxonomy for designing compressed domain computer vision solutions driven by the architecture and weights compatibility with an available spatio-temporal computer vision processor. The potential of the proposed taxonomy is demonstrated for the specific case of point cloud classification by designing novel compressed domain processors using the JPEG Pleno Point Cloud Coding standard under development and adaptations of the PointGrid classifier. Experimental results show that the designed compressed domain point cloud classification solutions can significantly outperform the spatial-temporal domain classification benchmarks when applied to the decompressed data, containing coding artifacts, and even surpass their performance when applied to the original uncompressed data.

## I. INTRODUCTION

Humans communicate with the surrounding world using their senses, with sight playing a major role. In recent decades, digital multimedia information, applications, and services have exponentially increased the human communication capabilities in key areas like personal communications, broadcasting, streaming, social networks, medicine, education, industry, cultural heritage, and virtual reality. The visual representation models have evolved from 2D pixel-based models with increasing resolution to sophisticated 3D models designed to offer the users highly realistic, immersive, and interactive experiences [1]. In this context, multimedia coding and compression technologies have been paramount in limiting the required transmission and storage resources for the target qualities, thus allowing multimedia data to reach more users and consumption conditions. The explosion of multimedia for human consumption through visualization was enabled by a number of very popular multimedia coding standards which offered user interoperability, thus enabling large scale adoption and deployment. In this context, it is fair to highlight the JPEG image coding standards [2]-[3] and the MPEG video (and audio) coding standards [4]-[6], which empower the codecs that have invaded our houses, offices, and pockets. For visual content, notably images and video, the conventional coding approaches rely on transforms to exploit the spatial redundancy, temporal differences and motion compensation to exploit the redundancy, entropy coding to exploit the statistical redundancy, and quantization to exploit the perceptual redundancy or irrelevance and control the coding rate. Along with the exponential growth in multimedia information produced, shared, and consumed by humans for visualization, under very heterogeneous conditions, the recent years have also seen the explosion of multimedia data consumption by machines in multiple types of computer vision tasks, notably classification, segmentation, detection, and recognition, again in a growing number of application domains. Often, the machine computer vision (CV) tasks are performed on previously compressed multimedia content (not the original data), used for more efficient storage or transmission. These data must be first decompressed before CV processing, since the best available CV technologies process spatio-temporal data and not directly compressed data. This decompressed domain approach for CV tasks is affected by the negative impact of the compression artifacts inherent to the lossy decoded data, which depend on the used rate. In this paper, the term 'spatio-temporal' (ST) refers both to original as well as decompressed data, and includes multimedia modalities with and without temporal variation, e.g., images and static point clouds as well as video and dynamic point clouds.

In the last decade, the multimedia community was shaken by the surge of machine learning technologies, boosted by the emergence of deep learning (DL) algorithms, the large-scale availability of data, and the advances in computational platforms. In particular, DL-based technologies have taken over the CV field by achieving dominating performance for a wide range of tasks, sometimes even above human performance [7]-[11]. This success may be largely attributed to the adoption of deep convolutional neural networks (CNN) [12], trained to identify and extract useful features for a specified task, as opposed to extracting handcrafted features as before. More recently, attention mechanisms such as transformers have boosted impressive advances in CV tasks such as image classification, where the Vision Transformer [13], a pure transformer-based model without any CNN layer, has demonstrated better performance than the state-of-the-art CNN models. The idea of the attention mechanism is to determine the most important input pixels based on the relationships between pixels. As for images, transformers have also been designed for point clouds (PCs), notably the Point Cloud Transformer [14] and Point Transformer [15], both showing significant improvements compared to the state-of-the-art in PC classification and semantic segmentation tasks.

More recently, DL-based multimedia coding has shown its power and quickly matched or surpassed the compression performance of state-of-the-art conventional coding standards [16]-[23], which were the result of decades of research and development. This evidence has led both JPEG and MPEG, the major multimedia coding standardization bodies, to study the potential of DL-based coding for several multimedia modalities. JPEG has already launched two learning-based coding standards, notably JPEG AI for image coding [24] and JPEG Pleno Point Cloud Coding (PCC) [25] for static PCs, with impressive results. While the current version of the JPEG AI codec (still under development) achieves more than 30% rate reduction over the powerful Versatile Video Coding (VVC) standard in its Intra mode [26]-[27], the current version of the JPEG Pleno PCC codec [28]-[29] outperforms the recently developed MPEG Geometry-based PCC (G-PCC) standard for static PCs [30]. Current activities place video and light fields as the next modalities in line for the development of DL-based coding standards.

Before the emergence of DL tools in the multimedia landscape, coding and CV processing lived independently in the sense that CV processing would happen after decoding, naturally suffering from the coding artifacts introduced by compression, since the best 'features' used to represent the multimedia data for efficient compression and good quality human visualization were not the same features as for CV processing. A clear example may be given for images: with image coding commonly performed with the first JPEG standard (and other JPEG standards after, although not as popular), CV processing would be performed with specific hand-made image descriptors like Scale-Invariant Feature Transform (SIFT) and Speeded-Up Robust Features (SURF) since the JPEG DCT coefficients (the coding features of that time), although efficient for coding and human visualization, would not allow efficient CV tasks processing. Previous works in the literature have studied the impact of performing CV tasks on decompressed content after lossy compression, notably for PC classification [31] and face image recognition [32]. These works clearly show that the CV tasks' performance is negatively impacted by the coding artifacts, especially at lower rates, as could be expected.

The above paradigm has totally changed when DL-based models started offering very efficient solutions for multimedia coding, at a stage in which they already offered the best solutions for CV tasks processing.

The adoption of DL-based media coding opened a new and bold horizon, which is the target of the two emerging JPEG coding standards, i.e., JPEG AI and JPEG Pleno PCC. Building on the success of DL-based tools for multimedia representation targeting both human visualization and CV tasks for machine consumption, the scope of these future JPEG standards envisions for both images and PCs '*the creation of a learning-based coding standard for images/point clouds and associated attributes, offering a single-stream, compact compressed domain representation, supporting advanced flexible data access functionalities. The standard targets both interactive human visualization, with competitive compression efficiency compared to state of the art image/point cloud coding solutions in common use, and effective performance for processing and machine-related computer vision tasks*' [24]-[25]. An illustration of the JPEG

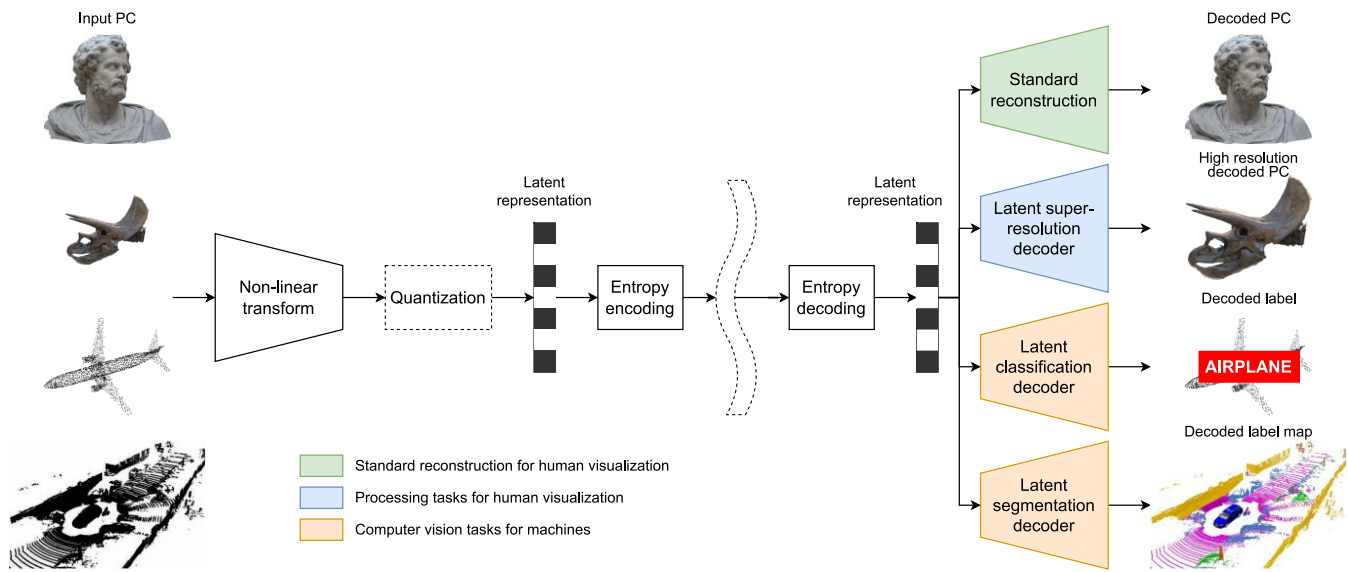

FIGURE 1. Illustration of the JPEG Pleno PCC unified representation framework for man and machine. Depending on the target use case, the same latent representation (coding stream) may be: (green) decoded to provide a regular/standard reconstruction; (blue) directly used by a DL-based super-resolution model to provide a higher resolution reconstruction; or (orange) directly processed to perform a CV task, such as classification or semantic segmentation, targeting machine consumption.

Pleno PCC unified representation framework for man and machine is shown in Fig. 1. The figure depicts some examples of the usage of a single *compressed representation* (coding stream), commonly known as *latent representation* or *latents*, for multiple use cases: the path in green corresponds to the past compression purpose, in which the latent representation is simply (standard) decoded to reconstruct the PC for human visualization, with a quality depending on the invested rate; the path in blue corresponds to an enhancement/processing task for human visualization, in this case super-resolution, with the goal to produce a higher resolution PC directly from the compressed latent representation,; the paths in orange correspond to two CV tasks, in this case classification and semantic segmentation, directly performed from the compressed latent representation, useful for machine consumption.

In this context, the development of DL-based compressed representations and standards, e.g., JPEG AI standard for images and JPEG PCC standard for PCs, must consider not only the usual Rate-Distortion (RD) performance to maximize the decoded quality for a target rate, aiming at human visualization, but also the CV task performance, e.g., detection, classification, recognition accuracy. This double approach may have a direct impact on the multimedia coding model, especially in the training phase and associated loss function, since they define the model optimization strategy and thus its performance. While in practice the model loss function may consider/combine both RD and CV task accuracy metrics (although only those differentiable), thus having the CV task processing directly impacting the coding model parameters, what this paper shows for PCs is that compressed domain CV processing is more efficient than decompressed domain CV processing, even if the coding model only considers the RD component in the loss function used for training. This is extremely important because it allows the easier adoption and deployment of this type of DL-based coding solutions for human visualization applications since they offer the most efficient solutions both for coding and CV processing, without having to accept any penalty in compression performance to reach some performance trade-off between the two types of purposes.

In summary, DL-based multimedia representation technologies will not only bring additional compression performance but shall also enable a common, compressed representation for multimedia information, effectively serving both man and machine, thus empowering decoding for human visualization and featuring processing for machine consumption from the same compressed stream [33]-[38]. Performing CV tasks on the compressed domain representation (as opposed to the current decompressed ST domain) has two major advantages: i) increased accuracy, by using features extracted from the original media data and not from its lossy decompressed version; and ii) reduced complexity, since compressed domain CV tasks do not require prior decoding and fully extracting features before processing This integrated vision is a breakthrough for multimedia content representation since it unites man and machine multimedia representation into a single, common framework. In the future, CV tasks may be performed over the *original ST* data or its *decompressed ST representation* (as nowadays) or, alternatively, directly over its learning-based *compressed representation*, containing the features extracted by appropriately trained DL models, depending on the application scenarios.

In this context, a compressed domain CV processor may be either designed from scratch or adapted from an available ST

domain CV processor to allow it to directly process compressed streams. The latter approach has the potential to offer some desirable compatibility between the ST and compressed processing domains, e.g., by using partly common CV processor model architectures and weights, thus reducing the overall complexity, e.g., avoiding the memory footprint increase required by independent models. For this reason, the compatibility-driven design approach for compressed domain CV processing is adopted and evaluated in this paper.

Since there are multiple ways to design a compatibility-driven approach, this paper initially proposes a taxonomy for the various possible alternatives for adapting an existing ST domain CV processor to directly process the compressed domain representation generated by a DL-based media codec, which may also be decoded for human visualization. The design configurations defined by the various taxonomy branches offer different trade-offs in terms of complexity, CV task performance, and compatibility between the compressed and ST domain CV processors. Thus, the proposed taxonomy, the first of its kind, may guide the design of compressed domain CV processors, subject to constraints related to specific CV tasks, application domains, or hardware requirements.

To demonstrate the potential of the proposed taxonomy, this paper then focuses on a specific visual modality and CV task, notably PCs and classification, a very attractive combination due to its practical relevance for several emerging applications, and designs novel compressed domain processors. PCs are a powerful 3D visual representation model which is becoming very popular for immersive and realistic human visualization experiences, e.g., virtual reality, as well as CV tasks, e.g., in autonomous vehicles. A PC consists of a set of points in the 3D space represented by their coordinates (x, y, z), known as the PC geometry. A PC can offer a realistic object/scene representation by using a massive number of points to model the surfaces, which may have associated attributes such as color or normal vectors.

As for the established 2D visual modalities, e.g., images and video, it is expected that PC data will be commonly available using a compressed format since the increasing resolution and the large amount of content will not be compatible with (huge) non-compressed storage or transmission resources. In the image world, most sensors do not even make available the raw, non-compressed version anymore and only a compressed version, commonly using the JPEG (compressed) format, is made available.

Assuming PCs will be mostly available in the compressed domain, whenever some CV task has to be performed, e.g., classification, segmentation, detection or recognition, it is natural that it is performed in the compressed domain, especially if better CV performance and lower complexity are achieved as demonstrated in this paper. Examples of important PC classification real-world applications that usually deal with a very large amount of PC data and can benefit from this paradigm are (see also Fig. 1):

1. **Geographical Information Systems** where compressed PCs (to save storage and transmission resources) may be segmented and classified depending on the region features, e.g., urban versus non-urban, different types of buildings in urban areas, and different types of urban artifacts.
2. **Autonomous vehicles** where the compressed PCs (to allow faster processing by transmitting less data in the pipeline) corresponding to objects around the street may have to be segmented and classified, e.g., isolated person, group of persons, tree, traffic sign, red light pole, seating bench, etc. Moreover, the persons may have to be classified depending on the motion, as standing, walking, running, etc., which are critical decisions to drive the autonomous vehicles' behavior, thus needing a high level of accuracy.
3. **Cultural heritage** where compressed PCs (to save storage and transmission resources) corresponding to different types of artifacts, e.g., sculptures, columns, craftsmanship, etc., may have to be classified in terms of type and style.
4. **Shopping** where compressed PCs (to save storage and transmission resources) in a selling website corresponding to different purposes, e.g., clothes, kitchen, office, etc., may have to be classified in terms of type or style.

To show the potential of compressed domain (static) PC classification, two key state-of-the-art DL-based tools have been selected, one for PC coding and another for PC classification, with the target to develop taxonomy-driven compressed domain PC classification solutions. Due to their performance and maturity, the selected solutions are the JPEG Pleno PCC standard codec [28] (at this date reaching final stage) and the PointGrid PC classifier [39].

Building on these two key components, this paper will design, assess, and compare several compressed domain (static) PC classification solutions, guided by the proposed taxonomy, with the following goals/challenges:
1. Offering a classification accuracy that is equal or better than the decompressed domain PC classification performance.
2. Offering a complexity equal or lower than the decompressed domain PC classification complexity.
3. Obtaining the previous benefits regarding PC classification using a compressed domain representation that does not penalize the RD performance. This means that the PC classification benefits are obtained at no cost in compression performance for human visualization, what is critical to avoid jeopardizing its deployment for applications in which the main focus is human visualization.

The obtained classification performance results show that compressed domain PC classification may outperform spatial domain (original and decompressed) static PC classification with a reduced complexity, especially for the lower rates, while also offering synergies and some degree of compatibility between the spatial and compressed domain classification models. In summary, as far as the authors know, this paper proposes the first taxonomy for the adaptation of DL-based ST multimedia CV processors to the compressed domain and validates the power and benefits of DL-based compressed domain classification for static PCs by designing novel compressed domain processors.

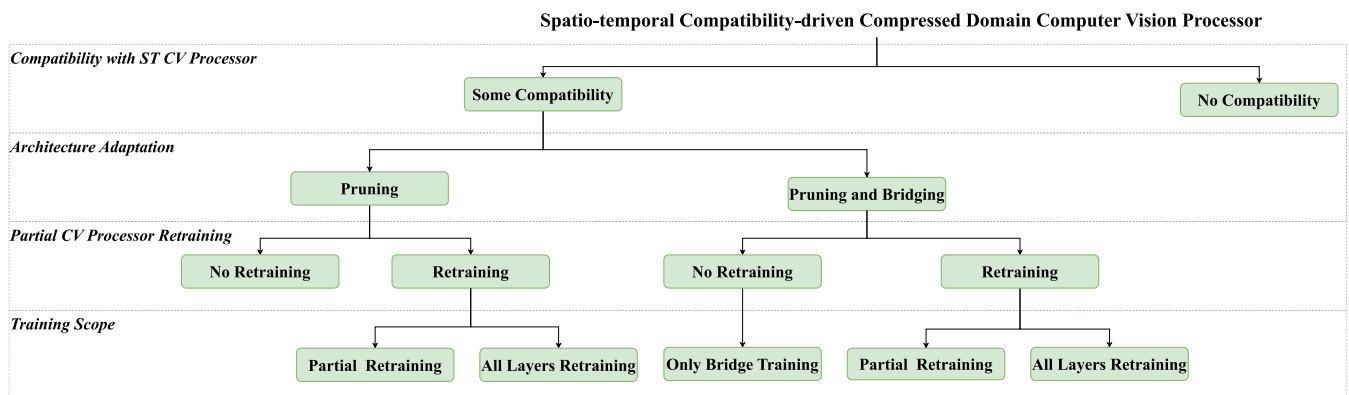

**FIGURE 2.** Overview of the proposed taxonomy for ST compatibility-driven compressed domain CV processing.

This paper is organized as follows: after this introduction, Section 2 proposes a taxonomy for the adaptation of ST domain CV processors to directly process compressed domain information, while Section 3 describes the relevant pipelines for performing a CV task. Next, Section 4 presents the specific PC classification pipelines and the adopted DL-based PC codec and PC classifier, while Section 5 presents the compressed domain PC classification solutions, designed according to the proposed taxonomy. Section 6 presents and discusses the experimental results and, finally, Section 7 concludes the paper and presents future work.

## II. TAXONOMY FOR COMPRESSED DOMAIN COMPUTER VISION PROCESSING

A taxonomy is a scheme of classification that allows to organize, structure and abstract the 'entities/solutions' in a specific field (in this case ST compatibility-driven compressed domain CV processing) with two main benefits: i) regarding the present, it makes it easier to discuss, analyze and compare the alternative solutions and abstract deeper relations, allowing a more profound knowledge and comprehension of the full landscape, notably the strengths and weaknesses of each solution; and ii) regarding the future, it makes it easier to understand the most promising design directions and their implications, not in an isolated way but rather organized by a taxonomic framework, e.g. allowing to identify features, strengths and weaknesses inherited from taxonomy parent and peer nodes.

Since many different design approaches may be used, this section proposes the first comprehensive taxonomy for the design of ST compatibility-driven compressed domain CV processing solutions, targeting the consumption of multimedia signals by both man and machine. This taxonomy will focus on compressed domain CV processing solutions offering some degree of compatibility with available ST CV processing solutions, since this is often a critical requirement to reduce complexity and exploit synergies between the compressed and ST CV processing domains, especially in scenarios where both compressed and decompressed content must be processed or when an existing computer vision processor with top performance has already been deployed. If no compatibility is required, the design of a DL-based compressed domain CV processor may be rather similar to the design (from scratch) of any other DL-based model, naturally considering the specific task at hand. While this approach has the potential to improve the task performance, this would come at the cost of higher memory requirements and also design effort, in comparison with a compatible approach.

The proposed taxonomy is agnostic to the specific compressed domain conditions, notably the media modality, the DL-based codec, the CV task and the ST DL-based processor. According to the proposed taxonomy, ST compatibility-driven compressed domain CV processors may follow rather different design approaches, each offering specific features and trade-offs, which allow to fulfill different requirements and thus effectively serve different use cases.

### A. TAXONOMY FOR COMPRESSED DOMAIN CV PROCESSING

The proposed taxonomy, represented in Fig. 2, is hierarchically organized according to four dimensions, which correspond to the horizontal layers shown in the figure. Within each dimension, different classes and branches in the taxonomy originate alternative designs, which define the main features and trade-offs differentiating the compressed domain CV processing configurations. The proposed taxonomy considers six branches which differ on the design solutions adopted for the four adopted taxonomy dimensions described in the following. This is graphically shown in Fig. 2 which offers a detailed overview of the proposed taxonomy for ST compatibility-driven compressed domain CV processing.

The taxonomy dimensions, organized in a top-down approach, i.e., from more fundamental to more specific options, are:

1. *Compatibility with ST CV Processor* – Refers to the key requirement of ensuring (or not) some compatibility between the compressed and ST CV processing domains. Two classes are defined with major implications on the design process:
   - *Some compatibility* – The development of the compressed domain CV processor is based on an available ST CV processor to offer some degree of compatibility between the compressed and ST domains, ideally with reduced complexity. This may happen either by sharing DL architecture layers and/or using common weights in the DL model layers. The proposed taxonomy is essentially about this compatibility-driven design and offers a way to organize its alternatives.
   - *No compatibility* – No compatibility requirements are considered at all and thus the compressed domain CV processor is designed from scratch, independently from any legacy ST CV processor. Since this approach is less constrained, it allows greater design flexibility with possible gains in the accuracy performance for the target CV task. The proposed taxonomy does not detail this branch since 'some compatibility' is the key design requirement in this paper.

2. *Architecture Adaptation* – This dimension regards how the architecture of the selected ST CV processor adapts to the input compressed domain stream. Since the starting CV processor is a DL-based model, often based on a CNN with several layers, the adaptation of the selected ST CV processor requires dimensionality matching between the input compressed data stream and the expected compressed domain CV processor input, which may be obtained in two main ways:
   - *Pruning* – The compressed domain CV processor is adapted to the compressed stream by simply pruning/removing some of the top layers (those closer to the input data) of the original ST CV processor, resulting in the so-called *Partial CV processor*.
   - *Pruning and Bridging* – The compressed domain CV processor uses an adaptation *Bridge* that is included between the compressed latent representation and the Partial CV processor. The Bridge is a DL model that acts as a preprocessing extension to the Partial CV processor. It is designed and trained to adapt the compressed domain latent representation of the multimedia signal (i.e., the output of the DL-based encoder) to the Partial CV processor, thus facilitating its task. The use of a Bridge has the potential to improve the performance of the compressed domain CV processor, at the cost of increased model complexity and training.

   It is important to note that the architecture of the compressed domain CV processor must guarantee the data volume matching, i.e., the compressed data stream exactly matches the dimensions required by the first layer of the Partial CV Processor. In some situations, depending on the used DL-based media codec and ST CV processor, this can be achieved by carefully pruning the original CV processor model in a way that ensures that the first layer of the Partial CV Processor uses a volume matching the dimensions of the latent representation generated by the DL-based codec. Otherwise, the use of a Bridge that ensures this so-called *data volume matching constraint* by adapting the dimensions of the latents' data volume is mandatory.

3. *Partial CV Processor Retraining* – This dimension refers to whether or not the Partial CV processor is retrained, with obvious implications on the degree of compatibility between the compressed domain and ST CV processors. The following two classes are defined:
   - *No Retraining* – The Partial CV processor inherits not just the architecture from the original ST CV processor but also the weights, i.e., no retraining is performed. While this class offers the highest level of compatibility, this may hinder the CV task performance.
   - *Retraining* – The selected Partial CV processor layers are retrained to further optimize their weights; here, reduced compatibility is accepted to potentially boost the CV task performance.

   While retraining the Partial CV processor limits the degree of compatibility between the starting ST and compressed domain CV processors, it may be important to improve the CV task performance, especially considering that the original weights were obtained for a CV processor version likely trained with uncompressed ST domain signals.

4. *Training Scope* – This dimension refers to the scope of the training/retraining process for the compressed domain CV processor. The relevant classes for this dimension are:
   - *Partial Retraining* – Only some of the compressed domain CV processor layers inherited from the ST CV model are retrained, whereas the remaining layers directly reuse the already available weights.
   - *All Layers Retraining* – All layers of the compressed domain CV processor are retrained.

   Naturally, when a Bridge is used, all its layers must be trained.

For all considered taxonomy branches, increasing the training scope (number of retrained layers) reduces the compatibility between compressed domain and ST CV processing but may lead to a better compressed domain CV processor performance. The configurations defined by the proposed taxonomy precisely express this trade-off which must be resolved when addressing the specific use case requirements.

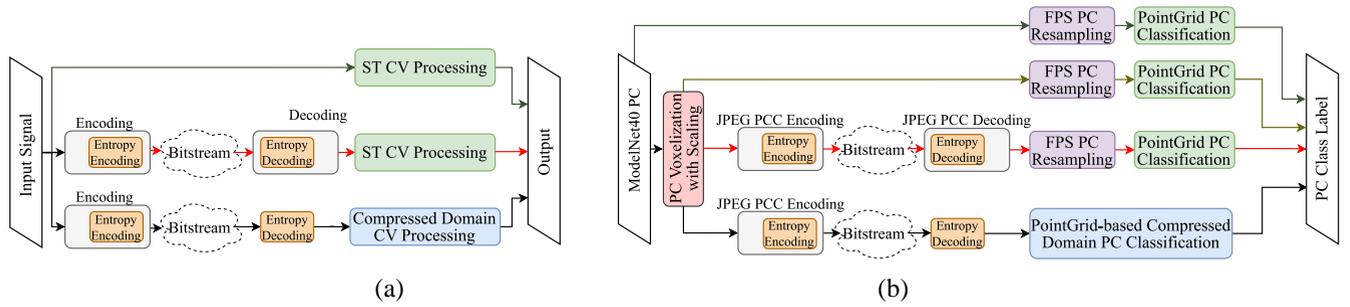

**FIGURE 3.** CV processing pipelines: a) generic in terms of multimedia modality, encoding/decoding and CV processing; b) specific for a modality (point clouds), encoding/decoding (DL-based JPEG PCC) and CV processing (classification with PointGrid).

### B. COMPATIBILITY WITH ST DOMAIN PROCESSORS

The design of the compressed domain CV processor is driven by a key requirement, which is the assurance of some level of compatibility with the selected ST domain processor. This compatibility involves two key aspects:

1. **Architecture compatibility** – Related to the number of layers in the compressed domain CV processor in common with the original ST CV processor architecture; in percentage, this compatibility may go from 0 to 100%.
2. **Weights compatibility** – Related to the number of weights in the compressed domain CV processor which are common with the original ST CV processor; in percentage, this compatibility may go from 0 to 100%.

These two types of compatibility are directly related to the possibility to reduce the storage (or transmission) required by the additional compressed domain CV processor, assuming that the selected ST domain CV processor is already available at the receiver. Full compatibility means that all the layers and weights of the compressed domain CV processor are inherited and reused. When no compatibility exists, the data describing the new architecture and all the weights defining the compressed domain CV model have to be stored or transmitted. It is important to note that typically the amount of data required to define a new model architecture is less than the data associated to the model weights. As a result, the impact of having no weights compatibility, even when full architecture compatibility is ensured, may be quite significant; this is especially relevant if transmission is required. Naturally, when there is no architecture compatibility, there is also no weights compatibility.

Adopting a compressed domain CV processor may allow achieving reductions in the overall complexity, here measured using the number of model parameters, in multiple ways:

1. First, by offering some degree of compatibility between the ST and compressed CV processing domains. Without any compatibility, two independent models would be required to accomplish the CV task in the ST or in the compressed domain. On the other hand, by keeping in common part of the CV processor model architectures and parameters for these domains, the overall complexity is reduced as this avoids the memory footprint increase of using independent solutions for the two domains.
2. Second, the compressed domain CV solutions avoid the complexity associated to the decoding process since the multimedia data, e.g., PCs, do not have to be decompressed.
3. Third, since compressed domain CV tasks do not require prior decoding, the process of feature extraction after decoding is also avoided as the features extracted at the encoder are directly used for the compressed domain CV processing.
4. Fourth, this paper designs compressed domain CV processors with a total number of parameters lower than the number of parameters for the original ST domain CV processor, thus imposing a hard limit on the complexity to guarantee that the developed compressed domain models are not more complex than the ST models. This is obtained, notably by controlling the Partial CV processor and Bridge architectures in terms of number and complexity of its layers.

All the compressed domain CV processor solutions designed in Section V and assessed in Section VI offer these complexity reductions.

### III. COMPUTER VISION PROCESSING PIPELINES

This section presents the main pipelines relevant for CV processing, which may include or not compression, depending on the use cases. The three pipelines presented in Fig. 3a will be used in later sections to study the performance of the designed compressed domain CV processor solutions, guided by the proposed taxonomy. The pipelines in Fig. 3a are (from top to bottom):

1. *Original Domain CV Processing Pipeline* – In this pipeline, no coding/compression is involved and thus the ST CV processor acts directly on the original (non-compressed) ST representation of the multimedia signal. Since no coding/compression is used, the CV processor performance is not impacted by any compression artifacts during the training or processing stages. However, if the use cases involve storage or transmission, large data resources for the (non-compressed) content may be required.
2. *Decompressed Domain CV Processing Pipeline* – In this pipeline, coding/compression and decoding/decompression occur. Therefore, the ST CV processor acts on a ST representation of the multimedia signal which is the result of compression and decompression. In this context, the ST CV processor performance is impacted by the compression artifacts caused by lossy coding the original multimedia signal. In this pipeline, the computational complexity associated to CV processing must also include the complexity associated to the signal decompression.
3. *Compressed Domain CV Processing Pipeline* – In this pipeline, coding/compression happens but no decoding/decompression is performed and, thus, the CV processor acts directly on the compressed domain multimedia signal representation, with the previously outlined benefits.

The three presented generic pipelines can be used to study the CV processing performance for any specific modality, codec, CV task, and CV processor.

## IV. COMPRESSED DOMAIN POINT CLOUD CLASSIFICATION

The taxonomy presented in Section II, as well as the processing pipelines described in Section III, address all multimedia modalities and accommodates any DL-based codec and ST CV processor combination. Following the previous generic presentation, agnostic in terms of modality and CV task, the following sections will address a particular application scenario, which has recently gained significant practical relevance: DL-based compressed domain PC classification. This section will start by presenting the specific compressed domain PC classification pipelines, where two key technologies play a fundamental role: DL-based (static) PC coding and (static) PC classification. Therefore, the following subsections will present the selected PC coding solution – JPEG Pleno PCC Verification Model (VM) [28], labeled from now on as *JPEG PCC* – and the selected ST PC classification solution – *PointGrid* [39]. These models will be used as the building modules for the design of multiple ST compatibility-driven compressed domain PC classification solutions, based on the proposed taxonomy. Both the JPEG PCC codec and the PointGrid classifier are DL-based solutions using a voxel-based approach with integer representation. For the PC classification experiments, there are a few publicly available PC datasets which may be used, such as ModelNet40 [40] and ScanObjectNN [41]. Due to its popularity, this paper has adopted the ModelNet40 PC dataset.

### A. PC CLASSIFICATION PIPELINES
The specific pipelines for the classification of (static) PCs encoded with JPEG PCC, reusing the PointGrid classifier, are presented in Fig. 3b, thus allowing a direct comparison with the generic processing pipelines presented in Fig. 3a. Fig 3b presents (from top to bottom):
1. *Original Domain PC Classification Pipeline* – The voxel-based PointGrid classifier is directly applied to the original ModelNet40 PCs with floating-point representation.
2. *Voxelized Domain PC Classification Pipeline* – The voxel-based PointGrid classifier is applied to a voxelized version of the original floating-point ModelNet40 PCs using an integer 8-bit precision PC representation. This pipeline allows assessing the impact of voxelization when compared with the previous pipeline.
3. *Decompressed Domain PC Classification Pipeline* – The voxel-based PointGrid classifier is applied to the PCs obtained after encoding/compression and decoding/decompression using a conventional or DL-based PC codec. The decoded/decompressed version of the PC may include compression artifacts which impact the PC classification performance.
4. *Compressed Domain PC Classification Pipeline* – A compressed domain PC classifier, designed and developed according to the taxonomy proposed in this paper, is applied to the latents generated by the DL-based JPEG PCC encoder, corresponding to the representation of the input PC in the compressed domain.

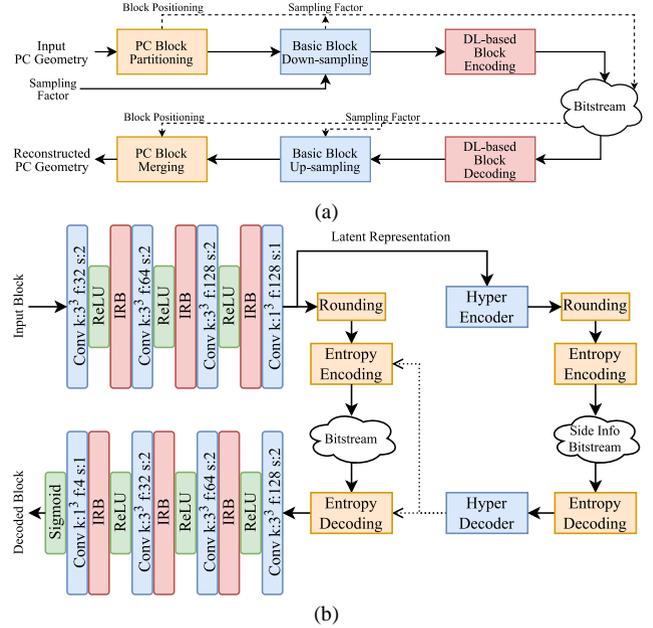

**FIGURE 4.** JPEG PCC codec: (a) high-level architecture; (b) DL-based block coding model architecture [24].

All the pipelines represented in Fig. 3b process the same input ModelNet40 PCs, each with 2048 points represented using floating-point values in a [-1, 1] range, eventually after voxelization and scaling. Moreover, all non-compressed domain pipelines use a *PC Resampling* module to ensure that the PointGrid classifier always receives a PC with 1024 points, since the classifier was trained for these conditions, for which is offers the best classification performance [39]. The Farthest Point Sampling (FPS) algorithm [42] is used to resample the decoded PCs.

### B. DL-BASED POINT CLOUD GEOMETRY CODING: JPEG PLENO PCC

With the focus of this paper being DL-based compressed representation and processing, it is imperative for both the PC codec and the PC classifier to be DL-based solutions. Given that JPEG Pleno PCC will be the first learning-based PC coding standard, currently at the latter stages of development, and the JPEG vision contemplates the design of a common compressed representation for both man (e.g., regular visualization) and machine (e.g., CV tasks), the ideal choice for a DL-based PC geometry coding solution is the JPEG Pleno PCC Verification Model (VM) [28] considering its potential future adoption, deployments and impact; the VM is a codec that evolves by collaboration towards the final DL-based JPEG Pleno PCC standard. The selected JPEG PCC codec corresponds to the so-called *IT-DL-PCC-G codec* [29] proposed to JPEG by the authors of this paper. This codec is used in this paper to create the latent representation which will be used by the DL-based compressed domain PC geometry classifier.

The high-level JPEG PCC architecture is shown in Fig. 4a, and can be divided into three major modules (both encoder and decoder). The encoder proceeds as follows:

1. **PC Block Partitioning**: Firstly, JPEG PCC divides the input PC geometry into 3D blocks, corresponding to coding units that are independently coded, with a selected block size (BS).
2. **Basic Block Down-sampling**: Secondly, the blocks may be down-sampled, using an appropriate sampling factor (SF), in order to reduce their precision and densify them prior to coding. This (optional) down-sampling is a tool that allows achieving lower rates, and can offer compression gains for sparse PC content.
3. **DL-based Block Encoding**: Finally, the actual coding of each 3D block is performed via a DL model, which architecture is shown in Fig. 4b.

The DL coding model uses a voxel-based representation of the PC data, which consists of a 3D block of binary voxels, corresponding to either occupied (1) or empty (0) positions. The DL coding model follows a similar approach as a conventional transform-based codec, consisting of a transform, followed by quantization of the transform coefficients and then entropy coding. The non-linear transform is learned by an autoencoder, consisting of multiple 3D convolutional layers that successively reduce the dimensionality of the data until the bottleneck; furthermore, Inception-Resnet Blocks (IRBs) [43] are used to extract high quality features, by containing multiple convolutional layers in parallel with different filter support sizes. The autoencoder thus creates a rich latent representation associated to each block of the input PC at its bottleneck layer. The latent representation is then quantized, and an adaptive entropy coding is performed using a secondary (hyper) network, which is responsible for extracting information from the latent representation itself, generating a more accurate entropy model.

The decoder proceeds in a symmetrical manner to the encoder as follows:

1. **DL-based Block Decoder**: The latent representation is first transformed back into the (lossy) reconstructed PC blocks using the DL coding model.
2. **Basic Block Up-sampling**: Then, if down-sampling was performed at the encoder, the inverse up-sampling process is applied to restore the original precision. Optionally, a DL-based super-resolution stage may be also applied to recover some of the data lost with the down-sampling, namely increasing the number of occupied voxels and ultimately improving the reconstruction quality.
3. **PC Block Merging**: Finally, the 3D blocks are merged together to generate the reconstructed PC.

To train the DL coding model, a rate-distortion (RD) driven loss function is used. This loss function balances the reconstructed PC distortion (estimated by the Focal Loss [44]) and the coding rate (estimated by the entropy of the latent representation). A Lagrangean multiplier, $\lambda$, is used to control the RD trade-off for compression. This requires training independent models with different values of $\lambda$, in order to encode and decode PCs for each target compression ratio [28]. The training of the models for each of the target compression ratios follows a sequential training approach, as opposed to training each of the models independently. With sequential training, the training of the model for the highest rate (lowest value of $\lambda$) is done first. Then, each subsequent model is trained by initializing its weights with the weights' values from the previously trained model.

JPEG PCC outperforms G-PCC Octree for coding geometry of dense PCs, while for sparse PCs (such as the ModelNet40 PC dataset), G-PCC Octree is still more efficient. For more details on JPEG PCC, refer to [28].

### C. DL-BASED POINT CLOUD CLASSIFICATION: POINTGRID CLASSIFIER

Considering the previous codec choice, namely JPEG PCC, selecting a PC classifier became more constrained since the way the PC data are represented and processed by both the codec and classifier must match. Since the JPEG PCC codec uses a voxel-based representation, the chosen PC classifier must also use a voxel-based representation instead of a point-based representation as used by the Point Cloud Transformer [14] and the Point Transformer [15]. This lead to the selection of the state-of-the-art PointGrid classifier [39], which was chosen due to its high classification performance compared to other ST voxel-based PC classification solutions with publicly available software, like VoxNet [45] and OctNet [46]. The PointGrid classifier model, represented in Fig. 5, consists of nine CNN layers, each using a LeakyReLU activation followed by batch normalization, and three fully connected (FC) layers, in a total of 10,492,072 weights. The PointGrid classifier was trained using the cross-entropy loss function between the ground truth and predicted classes of the ModelNet40 PC dataset. During training, all ModelNet40 PCs were resampled from 2048 to 1024 points. In this paper, the PointGrid model is applied to blocks of size 32×32×32 and four points per cell, since these parameters were reported to achieve the best classification performance for the ModelNet40 dataset [40]. Like most classifiers, the main limitation of the PointGrid classifier is that it processes input blocks with a fixed predefined size. This means that it may be necessary to use the intermediate Bridge model to perform the block size adaptation between the codec and the classifier, in addition to the latent representation adaptation.

## V. TAXONOMY-DRIVEN COMPRESSED DOMAIN POINT CLOUD CLASSIFICATION SOLUTIONS DESIGN

This section presents the design process for the compressed domain PC classification solutions studied in this paper, driven by the proposed taxonomy. The designed solutions process the latents from the JPEG PCC codec using a compressed domain classifier with different levels of compatibility with the originally selected ST domain PC classifier, i.e., PointGrid. Six solutions are proposed in the next subsection, corresponding to representative taxonomy branches and offering distinctive compatibility levels and architectural features. The design and training of these solutions are described in Subsections V.B and V.C, respectively. The performance of the designed solutions will be assessed in Section VI.

### A. SELECTED COMPRESSED DOMAIN PC CLASSIFIERS DEFINITION

This subsection proposes six different configurations for the compressed domain PC classifier, driven by the proposed taxonomy. The selection of these solutions was based on two main criteria: i) covering all the design approaches defined in the taxonomy; and ii) offering distinct compatibility levels between the compressed and ST domain CV processors to address different use cases.

Following the first criterion, the selected solutions are classified according to the two main design classes defined in the *Architecture Adaptation* dimension of the taxonomy for compatibility driven compressed domain classifiers (see Fig. 2): i) Pruning where the compressed domain classifier is composed only by a number of layers inherited from the ST DL model, referred as Partial Classifier; and ii) Pruning and Bridging where the compressed domain classifier uses a Bridge before the Partial Classifier to adapt the JPEG PCC latents to the Partial Classifier.

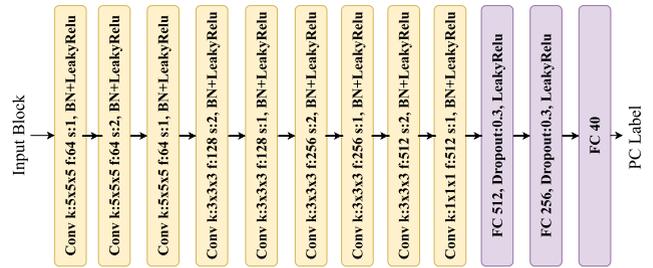

**FIGURE 5.** PointGrid classifier architecture [39].

For each of these two classes, three solutions are designed to offer distinct compatibility levels between the compressed and the ST domain CV processors:

- *High Compatibility (or HighComp)* – The percentage of compatible (meaning the same) weights between the ST and compressed domain classifiers is higher than 90%. This highest compatibility level is offered, both in terms of architecture and weights, by reusing Partial Classifier layers and corresponding weights from the original PointGrid classifier. In the proposed taxonomy (see Fig. 2), this corresponds to the "No Retraining" option for both the "Pruning" and "Pruning and Bridging" branches.
- *Medium Compatibility (or MediumComp)* – The percentage of compatible weights is medium, meaning higher than 30% and lower than 90%. A medium compatibility level is offered by still using the Partial Classifier architecture, but without fully reusing the weights, thus implying that some layers must be retrained using compressed domain data.
- *Low Compatibility (or LowComp)* – The percentage of compatible weights is lower than 30%. A low compatibility level is offered by retraining most of the Partial Classifier layers. The new weights for the retrained layers of the compressed domain PC classifiers need to be stored or transmitted as well as the Bridge weights.

The medium and low compatibility solutions fall into the "Partial Retraining" option for the" Pruning" and "Pruning and Bridging" branches (see Fig. 2). The "All Layers Retraining" option would correspond to 0% weights compatibility. The design process for these six proposed compressed domain PC classifier solutions is described in detail in the next subsection.

### B. SELECTED COMPRESSED DOMAIN PC CLASSIFIERS DESIGN

The proposed taxonomy guides the design of the compressed domain PC classifiers according to the hierarchy of design dimensions shown in Fig. 2. With this structured approach, the two main questions that guide the design of the compressed domain PC classifiers are:

1. How many of the PointGrid classifier bottom layers (closer to the output) shall be kept in the Partial Classifier.

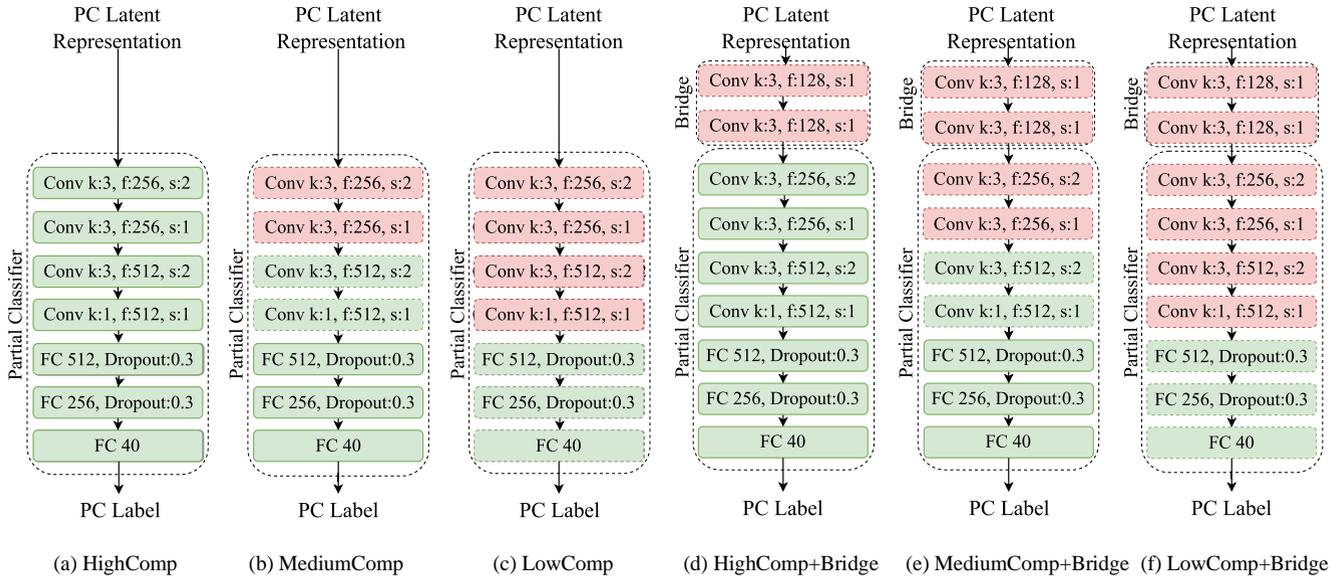

FIGURE 6. Architectures for the selected compressed domain PC classifier solutions. The colors for the layers represent the corresponding compatibility: Green – compatible weights, reused from the original PointGrid model; Red – not compatible (trained/retrained) weights.

TABLE I
ARCHITECTURE AND WEIGHTS COMPATIBILITY FOR THE SELECTED COMPRESSED DOMAIN PC CLASSIFIER SOLUTIONS

| Compressed domain PC classification solution | Number of layers | Architecture compatibility (in layers) | Total number of weights | Weights compatibility | Layers with weight compatibility | Number of new weights |
|---|---|---|---|---|---|---|
| HighComp | 7 | 7 (100%) | 8,699,176 | 8,699,176 (100%) | 4CNN+3FC | No retraining |
| MedComp |   |   |   | 6,043,432 (69.47%) | 2CNN+3FC | 2,655,744 |
| LowComp |   |   |   | 2,239,272 (25.74%) | 3FC | 6,459,904 |
| HighComp + Bridge | 9 | 7/9 layers (77,8%) | 9,584,168 | 8,699,176 (90.77%) | 4CNN+3FC | 884,992 |
| MedComp + Bridge |   |   |   | 6,043,432 (63.06%) | 2CNN+3FC | 3,540,736 |
| LowComp + Bridge |   |   |   | 2,239,272 (23.36%) | 3FC | 7,344,896 |

2. Is a Bridge needed and, if yes, what shall its architecture be, notably in terms of type and number of layers, kernel size, stride, and activation function.

As previously mentioned, the key design constraint for the compressed domain PC classifier is the matching of the data volume dimensions between the JPEG PCC codec and the compressed domain classifier. Depending on the design option, i.e., *Pruning* or *Pruning and Bridging*, this can be ensured by:

- *Pruning* – Selecting a pruning point for the PointGrid model so that the input to the first layer of the Partial Classifier matches the output JPEG PCC latent data volume.
- *Pruning and Bridging* – Designing the Bridge so that its model performs the adaptation between the JPEG PCC latent volume dimension and the Partial Classifier input. This can be achieved by carefully designing the architecture, filters, and strides for the down/up sampling layers using a CNN-based Bridge. To better illustrate the possible designs, this paper uses a combination of PC codec and compressed domain PC classifier where the matching of the data volume dimensions may be achieved without using a Bridge. In practice, this means that the fulfilment of the data volume constraint may be performed by only using the Partial Classifier. It is important to note that this design is not always possible, notably when there is no available layer of the ST domain CV processor for which the input block dimensions match the PC latent representation dimensions.

A second design constraint is, naturally, the complexity of the compressed domain PC classifier. This paper targets compressed domain PC classifiers with a total number of weights lower than the number for the original ST domain PointGrid classifier (10,492,072). This is a relevant requirement for the practical deployment of compressed domain CV processors.

In order to simplify the presentation and allow for a fairer comparison between the six proposed design solutions, all these solutions use a common architecture for the Partial Classifier and the Bridge (when used). This can be easily seen in Fig. 6, which shows the architecture details for the six selected solutions. For better distinction, the solutions using Pruning and Bridging include "+ Bridge" in their name. The

Partial Classifier and Bridge models have the following design features:

- **Partial Classifier** – Corresponds to a pruned version of the PointGrid model, so it inherits the architecture for the non-pruned layers (shown in Fig. 5). To directly fulfil the data volume matching constraint, pruning is performed at the fifth layer from top, meaning that the bottom seven PointGrid model layers (4 convolutional and 3 FC) are kept. This pruning point was selected since it ensures that the first layer of the Partial Classifier receives a volume of size 8×8×8×128, thus matching the dimensions of the most efficient JPEG PCC latent representation, as discussed later in Section VI. In summary, the Partial Classifier design uses (see details in Fig. 6):
  o Number of layers: 7
  o Type of layers:
    - 4 convolutional layers
    - 3 FC layers:
  o Number of weights: 8,699,176.

The direct use of the Partial Classifier as compressed domain classifier ensures 100% architecture compatibility. However, the level of weight compatibility depends on the training of the corresponding compressed domain PC classification solution, which will be described in Subsection V.C.

- **Bridge** – Corresponds to a new DL model that is designed and trained to better adapt the JPEG PCC latents produced by the JPEG PCC encoder to the Partial Classifier. To address the data volume matching constraint, and considering the previously chosen Partial Classifier design, the Bridge design requires the use of layers with stride 1 and 128 filters, so that the volume dimensions of the processed block are not altered (and thus matching). Furthermore, the complexity constraint limits the number of layers leading to the following proposed Bridge architecture (see details in Fig. 6):
  o Number of layers: 2
  o Type of layers:
    - 2 convolutional layers (3×3×3 kernel, 128 filters, stride 1)
    - LeakyReLU activation function.
  o Number of weights: 884,992.

Experimental results have shown that this Bridge model provides the best trade-off between complexity and classification performance, while ensuring that the compressed domain PC classifier total number of weights falls below the 10,492,072 weights of the original PointGrid model. The training of the Bridge will be described in the next subsection.

Table I presents a summary of the main design features as well as a detailed comparison of the number of weights and the compatibility levels for the designed compressed domain PC classification solutions. It may be observed that the weights compatibility ranges between 23.36% and 100% and the architecture compatibility between 77,8% and 100%. For all designs, the total number of weights is lower than for the ST domain PointGrid classifier (10,492,072).

### C. COMPRESSED DOMAIN PC CLASSIFIER TRAINING

After designing the DL models for the Partial Classifier and Bridge, their training/retraining process must be defined, depending on the selected compressed domain PC classification solutions, notably the target compatibility level. Since the Bridge layers are always new, the compatibility level for the compressed domain PC classifier is directly dependent on the number of compatible (i.e., not retrained) Partial Classifier layers, as shown in Table I. As expected, the training procedure depends on the two architecture adaptation options:

- *Pruning* – Only requires the retraining of the non-compatible (top) layers of the Partial Classifier as follows:
  o The Partial Classifier is initialized with the original PointGrid model weights.
  o The compatible layers weights (represented in green in Fig. 6) are frozen during retraining.

  The non-compatible layers weights (represented in red in Fig. 6) are retrained via fine tuning. Unlike training using random initialization, fine tuning considers the information learned from the original training in the ST domain, akin to a transfer learning process.

- *Pruning and Bridging* – Requires training the Bridge layers, followed by retraining the non-compatible layers of the Partial Classifier**:**
  o The Partial Classifier is initialized with the original PointGrid model weights.
  o The Bridge is trained considering a random initialization, while the Partial Classifier weights are frozen.
  o After training the Bridge, its weights are frozen, and the non-compatible layers of the Partial Classifier are retrained via fine tuning.

Differently from ST domain PC classification, the proposed compressed domain PC classifiers have to be trained with compressed data. This means that the dataset used for training each of the compressed domain PC classifier models is composed not by the original PCs in the training dataset, but their JPEG PCC latent representation. For this purpose, the training dataset PCs were compressed using each of the six JPEG PCC models, each trained for a different compression ratio (i.e., value of λ) to cover a range of relevant rates/qualities. Due to the nature of the learning-based codecs, the latent representations generated by the coding models targeting different compression ratios tend to vary significantly, implying that the latents in the same position may represent very different PC features. As such, it is required to train six classifier models (one for each target compression ratio) for each of the designed compressed domain PC solutions. While this does increase the memory footprint, it is a fundamental requirement, which in turn is dependent on the selected learning-based codec. To train the various models for each proposed compressed domain PC classifier solution, the same sequential training approach previously explained for the training of the JPEG PCC coding models was followed. This approach significantly reduces the overall training time and allows each compressed domain classifier model to build on the learning process for the models trained for higher rates.

Having designed the six compressed domain classifier solutions offering different architectural options and levels of compatibility with the ST domain, the next section presents their experimental validation and performance assessment.

## VI. PERFORMANCE ASSESSMENT

This section presents the experimental setup and performance assessment for the proposed compressed domain PC classification solutions. These solutions are compared with ST domain (original, voxelized and decompressed) PC classification solutions, using the four PC classification pipelines presented in Section IV. After describing the test conditions and experimental setup, Subsection VI.B reports and analyses the RD performance for the DL-based JPEG PCC codec and the benchmark G-PCC Octree standard. After Subsection VI.C reports and analyses the classification performances for the various ST domain classification pipelines, which will act as performance benchmarks for the proposed compressed domain PC classification solutions in Subsection VI.D.

### A. TEST CONDITIONS AND EXPERIMENTAL SETUP

This subsection presents the test conditions and the experimental setup used for performance assessment.

**Dataset** – All training and tests PCs are from the Modelnet40 dataset, which is the most used dataset for PC classification [40]. The ModelNet40 PC dataset is composed of 12,308 PCs divided into 40 classes, corresponding to objects represented as 3D CAD models. The PCs are grouped into three subsets, used for training, validation, and testing, with 9,840, 2,048, and 420 PCs, respectively. All PCs have floating-point precision, with coordinates' values ranging from -1 to 1, and a total of 2048 points per PC.

Regarding the training datasets, it is relevant to highlight:
- The training of the PointGrid PC classifier was performed as in [32], i.e., the original PCs from the ModelNet40 training dataset with floating point representation were used.
- The DL models for the JPEG PCC codec were trained according to the PCC Common Training and Test Conditions (CTTC) set by JPEG [47]. The training dataset was composed by 28 static PCs with varying features, notably in terms of resolution and sparsity. It is important to note that the codec was not trained using the same dataset used for the PointGrid classifier. Six values of λ (0.008, 0.004, 0.002, 0.001, 0.0005 and 0.00025) were used to train six JPEG PCC coding models, to reach different target rates/qualities.
- The training dataset for each compressed domain PC classifier models (one for each λ value) was composed by the PC latent representation for the ModelNet40 training dataset. This means that each PC was compressed using the JPEG PCC model trained for the corresponding compression ratio (i.e., λ value). Based on the RD performance, the JPEG PCC coding configurations were set to BS=64 and SF=4 as will be explained in the next subsection.

While it would be possible to train JPEG PCC using the same ModelNet40 dataset, which could potentially improve the RD performance, and perhaps even the compressed domain classification performance, the current JPEG PCC training dataset consists of a richer set of diverse PCs with more complex and realistic objects and scenes than the ModelNet40 dataset, which is more representative of the real-world content as targeted by the JPEG PCC coding standard. Using the ModelNet40 dataset to train the JPEG PCC codec would likely have a negative impact on its RD performance for real-world PC content, as shown in previous studies in the literature such as [48].

Regarding the test datasets, it is relevant to highlight:
- All tests performed in this paper, both for compression and classification performance assessment, used the geometry of the 420 PCs in the ModelNet40 test dataset.
- For the original and voxelized domains PC classification pipelines (see Fig. 5), the original and voxelized (with 8-bit precision) versions of the test PCs were classified using the PointGrid classifier.
- For the decompressed domain PC classification pipeline (see Fig. 5), the test PCs were encoded and decoded using the selected codec (JPEG PCC or G-PCC Octree) and then classified using the PointGrid classifier.
- For the compressed domain PC classifiers (see Fig. 5), the test PCs were encoded using the JPEG PCC codec and then the latent representation was used for classification using one of the designed compressed domain PC classification solutions.

**Training Hyperparameters** – The Adam optimizer was used to train the designed compressed domain PC classification solutions, with a batch size of 32. The cross-entropy loss function was used between the predicted and ground truth classes. For the compressed domain PC classifiers, early stopping with a patience of 50 epochs was used. The learning rate was set to $10^{-4}$, halving it whenever the classification accuracy over the validation dataset stopped increasing for 10 epochs.

**Compression Performance Metrics** – The RD performance for the ModelNet40 PC dataset compressed using the two selected geometry-only codecs, JPEG PCC and G-PCC Octree, was evaluated using the PSNR D1 metric, since this is the fidelity geometry quality metric recommended by both MPEG Common Test Conditions (CTC) [49] and JPEG CTTC [47]. The compressed rate was measured in bits per input point (bpp).

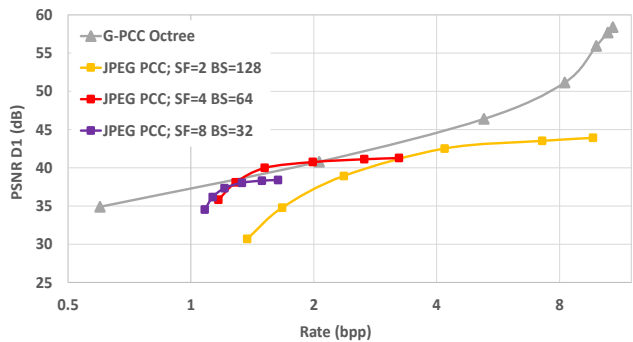

**FIGURE 7.** RD performance for G-PCC Octree and JPEG PCC using three relevant coding configurations.

The PSNR D1 metric is based on the Euclidean distance (error) between each pair of points of the reference PC, referred to as *A*, and the PC under evaluation (decoded), referred to as *B*. For each point in *B*, its associated point in *A* is determined using a nearest neighbor algorithm. Let $e_{B,A}$ be the arithmetic mean of the errors of all points in *B*. To ensure that the PSNR D1 metric is symmetric and invariant to which of the PC is used as reference, both $e_{B,A}$ and $e_{A,B}$ are used in the final PSNR D1 computation:

$$PSNR\ D1 = 10 \log \frac{3\ peak^2}{\max(e_{B,A}, e_{A,B})}, \quad (1)$$

where $peak = 2^{bit\ depth} - 1$ is the geometric resolution associated with the voxel-based PC representation, e.g., for a voxel bit depth of 10, *peak*=1023 [47].

**Classification Performance Metrics** – The classification performance for all tested PC classification pipelines was measured using the Top-1 and Top-5 metrics, since they are the most predominantly adopted metrics in the literature. Top-1 is the percentage of test examples for which the class with the highest probability exactly matches the ground truth. Top-5 is the percentage of test examples for which the ground truth is included in the 5 classes with the highest probabilities output by the classifier. Naturally, Top-5 is always equal to or higher than Top-1.

For compressed domain PC classification, the Bjontegaard Delta metric (BD)-Top-k is also used to compare the classification performance (versus rate) of two PC classification solutions. BD is a common metric to assess the improvement of one performance curve over another peer performance curve used as reference [50]; a positive BD-Top-k implies a Top-k improvement for the same rate and vice-versa. The BD metric is extensively used in the literature for comparing the RD performance of different codecs, e.g., for images and PCs.

### B. RD PERFORMANCE

This subsection reports and analyses the RD performance for the selected PC codecs, one learning-based and one conventional, i.e., JPEG PCC [29] and G-PCC Octree [30]. Analyzing the RD performance is important since the (lossy) compression artifacts impact the classification performance for the decompressed and compressed pipelines. Fig. 7 shows the average RD performance for the 420 test PCs of the ModelNet40 test dataset, for the two PC codecs. The G-PCC reference software, TMC13, version v14 was used under the configurations defined in the MPEG CTC [49]. For JPEG PCC, three different coding configurations are consideredfor the SF and BS parameter values: SF=2 BS=128, SF=4 BS=64, and SF=8 BS=32; the study of the compression performance for these configurations will allow to identify the most appropriate for the ModelNet40 test dataset, notably considering its sparsity. The results in Fig. 7 allow concluding:

- G-PCC Octree offers a better RD performance than JPEG PCC, notably for the higher rates, what is explained by the very high sparsity of the ModelNet40 PCs. However, this behavior is not the same for denser PCs coding, for which the JPEG PCC codec has a clear RD performance advantage [29].
- For JPEG PCC, the SF and BS configuration values have a noticeable impact on both the reconstruction quality and coding rate.
- Higher values for the JPEG PCC SF parameter allow to achieve a better RD performance for very low rates, while lower SF values allow improving the RD performance for higher rates.
- The JPEG PCC configuration using SF=4 and BS=64 offers the best overall RD performance trade-off for coding the selected ModelNet40 test dataset.

The next subsection presents the PC classification performance for the ST domain (original, voxelized, and decompressed) pipelines.

### C. SPATIAL TEMPORAL DOMAIN PC CLASSIFICATION

This subsection reports and analyses the PC classification performance for the three ST domain classification pipelines, i.e., original, voxelized and decompressed, described in Section IV and presented in Fig. 5. Fig. 8 presents the average classification accuracy as a function of the rate (in bpp), for the Top-1 and Top-5 classification metrics, for the adopted ModelNet40 PC test dataset. The results in Fig. 8 allow concluding:

- For the original and voxelized domains, PointGrid offers similar PC classification performance for both Top-1 and Top-5, meaning that the voxelization of the input PCs (to 8-bit) has no major impact.
- For the decompressed domain, JPEG PCC generally offers better classification performance than G-PCC Octree, notably for the SF=4 BS=64 and SF=8 BS=32 configurations, for both Top-1 and Top-5. Note that this does not follow the same relative behavior as for the compression performance shown in Fig. 7 (where G-PCC Octree shows better RD performance), meaning that JPEG PCC introduces compression artifacts on the reconstructed PC that do not seem to impact as much the PC classification performance as those introduced by G-PCC Octree.
- For the JPEG PCC codec, the classification performance depends significantly on the used coding configurations.
- The PC classification performance for the decompressed ModelNet40 PC test set follows closely the relative performance behavior observed for the RD performance.

Besides the objective RD performance and classification performances, human visual inspection has shown that the JPEG PCC (de)coded PCs with the SF=4 BS=64 configuration look visually better. As this paper targets both man and machine consumptions, this is the JPEG PCC coding configuration selected to assess the performance of all the designed compressed domain PC classification solutions, discussed in the next subsection.

### D. COMPRESSED DOMAIN PC CLASSIFICATION

This subsection reports the PC classification performance for the six compressed domain classification solutions designed and presented in Section V. Fig. 9 presents the average accuracy as a function of the rate (in bpp) for the Top-1 and Top-5 classification metrics for the adopted ModelNet40 PC test dataset. Table II presents a comparison between the compressed and decompressed domains PC classification curves using the BD-Top-k metric. The reference pipeline/configuration selected for the BD-Top-k computation is the Decompressed JPEG PCC with SF=4 BS=64 solution. These results allow concluding:
- When considering a HighComp solution without any retraining and without a Bridge, the classification results are catastrophically low (the BD losses reach 77.5% for Top-1 and 83% for Top-5). This is expected since there is no adaptation between the latents produced by the JPEG PCC codec and the latents expected by the designed HighComp classifier, thus demonstrating the need to design and train better solutions as described in Section V.
- On the other hand, all other designed compressed domain PC classification solutions offer considerably better classification performance than the decompressed domain classification solutions. This is most notable at lower rates, where decompressed domain PC classification tended to badly suffer from the heavier compression artifacts, whereas the compressed domain classification produces a more robust performance even at lower rates since latents extracted from the original PCs are used. Furthermore, some of the compressed domain PC classification solutions even present gains when compared with the PointGrid classifier applied to the original or voxelized PCs; this can be explained by considering that the JPEG PCC autoencoder layers generate features that

TABLE II
BD-TOP-1 AND BD-TOP-5 FOR THE SELECTED COMPRESSED DOMAIN PC CLASSIFICATION SOLUTIONS

REFERENCE: DECOMPRESSED DOMAIN USING JPEG PCC WITH SF=4 BS=64

| Selected compressed domain PC classification solutions | BD-Top-1 | BD-Top-5 |
|---|---|---|
| HighComp | -77.49 | -82.93 |
| MediumComp | 7.92 | 2.30 |
| LowComp | 7.14 | 2.11 |
| HighComp+Bridge | 8.75 | 2.98 |
| MediumComp+Bridge | 9.33 | 3.30 |
| LowComp+Bridge | 9.67 | 3.29 |

are richer than those generated by the pruned layers from the original PointGrid.
- The solutions using a Bridge have a PC classification performance that is always better than the corresponding versions only using a Partial Classifier (Pruning solutions). This is expected since the additional layers in the Bridge result in a deeper model with more layers trained specifically for PC classification based on the latent representation.
- For the designed compressed domain PC classification solutions, as the compatibility level increases, the PC classification performance decreases. This is easily observed for the solutions using a Bridge, where the best classification performance is achieved by the low compatibility solution (9.67 versus 8.75 for BD-Top-1 for the LowComp and HighComp compatibility solutions). This demonstrates the expected trade-off between compatibility and classification performance.
- The compatibility versus classification trade-off is not as evident for the solutions only using a Partial Classifier, since the LowComp and MediumComp solutions have approximately the same PC classification performance (the HighComp solution has extremely low PC classification results, as mentioned before).
- For BD-Top-5, the compressed domain PC classification performance follows the same relative performance behavior as for BD-Top-1, with more similar performances for all solutions since Top-5 is a 'more relaxed' classification metric.

The results clearly demonstrate that the design of the compressed domain PC classification solution has a strong influence on the related classification performance, even more than the influence of the quality/rate of the latent representation. The use of a Bridge (even with the simple architecture proposed for the designed solutions to limit the complexity) is a clear requirement for the improving PC classification performance. It is important to note that all the designed compressed domain PC classification solutions use a lower number of weights than the original PointGrid classifier.

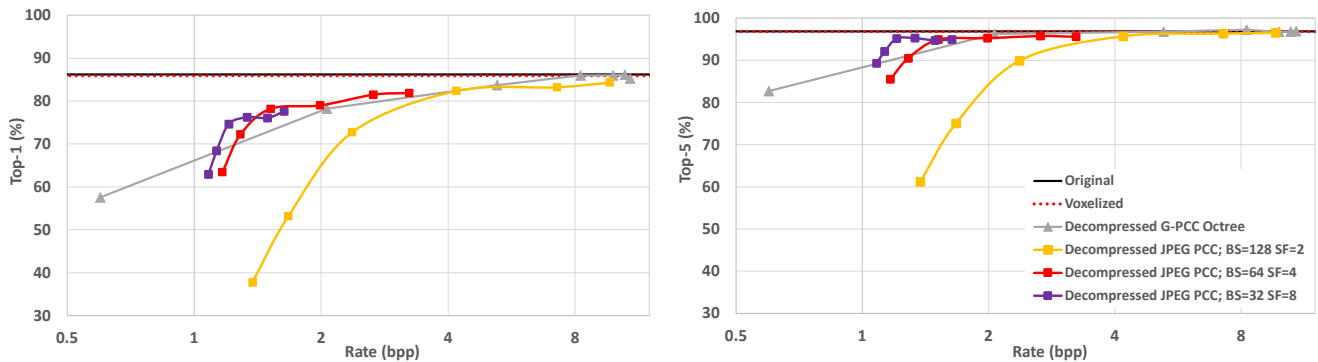

**FIGURE 8.** Top-1 (left) and Top-5 (right) classification performance for the adopted ST Domain PC classification pipelines.

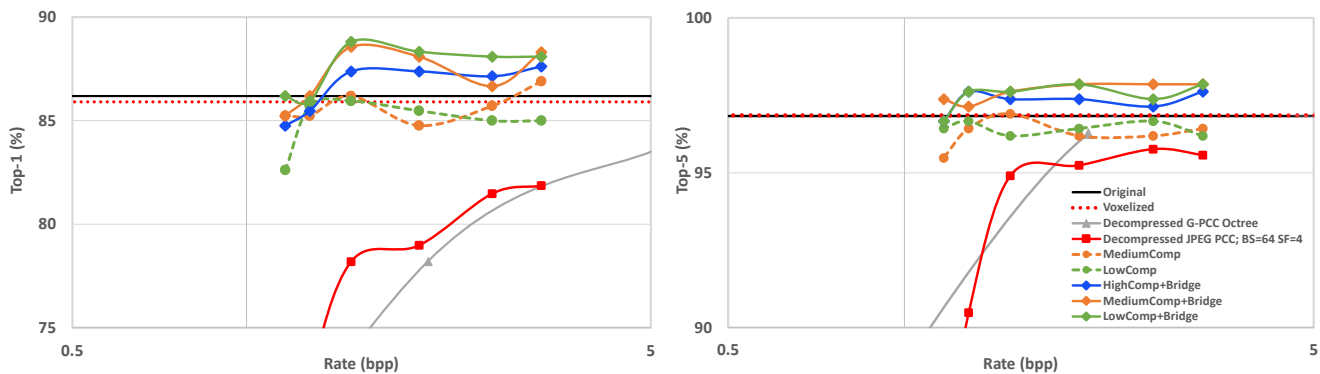

**FIGURE 9.** Top-1 (left) and Top-5 (right) classification performance for the adopted compressed domain PC classification pipelines. For all solutions involving JPEG PCC coding, SF=4 and BS=64 are used. Note that the results for the HighComp solution are not shown in the plots since they fall well below the shown performance range: from 0.7% to 1.2% for Top-1 and 8.3% to 15.95% for Top-5.

### E. PC CLASSIFICATION PERFORMANCE SENSITIVITY STUDY

Table III shows the results of a sensitivity study targeting to assess the performance of compressed domain PC classification solutions with progressively lower weights compatibility. The first half of the table corresponds to solutions using only a Partial Classifier and the second half to solutions also using a Bridge. The weights compatibility, shown in the second column, is varied by decreasing the number of Partial Classifier layers e with weights compatibility (i.e., the frozen layers), shown in column 3. The results allow concluding:

- There is a general tendency for increasing the PC classification performance as the level of compatibility decreases, although the variations may be small and not strictly monotonical.
- All the solutions with Bridge offer better classification performance than the solutions without Bridge.
- While the difference between the HighComp and the first MediumComp solution (shown in the first two rows of the table) is limited to the retraining of a single Partial Classifier CNN layer, the small variation of 100% to 89,82% in weights compatibility corresponds to a huge gain in PC classification performance (from a 77.49% loss into a 3.22% gain for BD-Top-1). This

TABLE III
BD-TOP-1 AND BD-TOP-5 FOR DIFFERENT COMPRESSED DOMAIN CLASSIFICATION SOLUTIONS

REFERENCE: DECOMPRESSED DOMAIN USING JPEG PCC WITH SF=4 BS=64

| Compressed domain PC classification solution | Weights compatibility | Layers with weight compatibility | BD Top-1 | BD Top-5 |
|---|---|---|---|---|
| HighComp | 100% | 7/7 | -77.49 | -82.93 |
| MediumComp | 89.82% | 6/7 (3CNN+3FC) | 3.22 | 0.85 |
|  | 69.47% | 5/7 (2CNN+3FC) | 7.92 | 2.30 |
|  | 28.77% | 4/7 (1CNN+3FC) | 7.96 | 2.22 |
| LowComp | 25.74% | 3/7 (3FC) | 7.14 | 2.11 |
|  | 0% | 0/7 | 7.93 | 2.22 |
| HighComp+Bridge | 90.77% | 7/9 (4CNN+3FC) | 8.75 | 2.98 |
| MediumComp + Bridge | 81.53% | 6/9 (3CNN+3FC) | 9.15 | 3.20 |
|  | 63.06% | 5/9 (2CNN+3FC) | 9.33 | 3.30 |
|  | 26.12% | 4/9 (1CNN+3FC) | 8.97 | 2.98 |
| LowComp+Bridge | 23.36% | 3/9 (3FC) | 9.67 | 3.29 |
|  | 0% | 0/9 | 9.19 | 2.86 |

is naturally due to the absence of any JPEG PCC latents adaption in the HighComp solution.
- Although training just the first Partial Classifier CNN layer, the MediumComp (3CNN+3FC) solution already offers better classification performance than

- the reference decompressed domain PC classification for BD-Top-1 and BD-Top-5.
- For both LowComp cases, with and without Bridge, when retraining all layers including the FC layers, the classification performance does not improve much. This may be explained by the fact that these layers are fundamentally responsible to aggregate the features and perform the final classification, and so they do not benefit from finetuning with compressed data.

Despite all the previous considerations being related to the particular conditions used in this performance assessment, the obtained results clearly validate the vision of compressed domain CV processing for the future JPEG PC coding standard. Moreover, given the generally good gains for all tested compressed domain solutions, the selection of the best compressed domain PC classification solution in terms of compatibility with the original ST PointGrid classifier may be based primarily on considerations related to exploiting weights or architecture compatibility associated with each particular use case.

## VII. CONCLUSIONS AND FUTURE WORK

This paper proposes a taxonomy, the first of its kind, for the design of ST compatibility-driven compressed domain CV processing solutions, targeting the consumption of multimedia signals by both man and machine using a single compressed stream. The constraints related to performing CV tasks in the compressed domain are presented considering the specific case of PC classification. To demonstrate the potential of the proposed taxonomy, the state-of-the-art DL-based JPEG PCC codec and the DL-based PointGrid PC classifier were selected. Driven by the proposed taxonomy, six compressed domain PC classification solutions were designed, covering different branches of the taxonomy, and offering varying compatibility levels with the original PointGrid classifier. The experimental PC classification results for four alternative classification pipelines (original, voxelized, decompressed and compressed domains) show that the compressed domain PC classification solutions offer the best classification performance, while achieving a reduction in the overall classifier model complexity. These results demonstrate the potential of the current JPEG vision to develop multimedia coding standards that simultaneously accommodate efficient fidelity-to-original coding and effective CV tasks processing, not only for PCs but also for images with JPEG AI.

Despite the positive results obtained, the chosen codec and classifier have some limitations. In future work, the constraint imposed by the computationally heavy voxel-based JPEG PCC representation may be overcome by adopting a sparse tensor representation, which works similarly to the voxel-based representation, but it only explicitly represents the occupied voxels and their coordinates; this makes it significantly lighter in terms of computational complexity and allows the use of point-based classifiers as well as voxel-based classifiers. As for the classifier, in this case PointGrid, its major limitation is the use of a fixed input block size, which may be overcome by designing an adaptive Bridge that is able to resize the compressed latent representation produced by the codec into the size expected by the Partial Classifier.

Future work will also follow the development of JPEG DL-based coding standards and develop architectures for other compressed domain CV tasks and modalities.

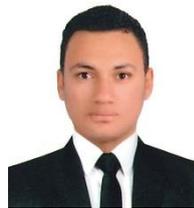

**ABDELRAHMAN SELEEM** (Student Member, IEEE) received the B.Sc. and M.Sc. degree in computer science from the Computer Science-Mathematics Department, Faculty of Science, South Valley University, Egypt in 2016 and 2021, respectively. From March 2022 to the present, he is a Ph.D. student at the Department of Electrical and Computer Engineering, Instituto Superior Técnico, University of Lisbon, Lisbon, Portugal, and a junior researcher at Instituto de Telecomunicações, Portugal. From April 2017 to August 2020, he worked as a Teaching Assistant at the Computer Science-Mathematics Department, Faculty of Science, South Valley University, Qena, Egypt. He worked as a Teaching Assistant and Lecturer Assistant from September 2020 to May 2021 and from June 2021 to February 2022, respectively, at the Computer Science Department, Faculty of Computers and Information, South Valley University, Qena, Egypt. He is actively contributing to the standardization efforts of JPEG on learning-based point cloud coding. His research interests include deep learning, computer vision, image processing, point cloud coding and analysis.

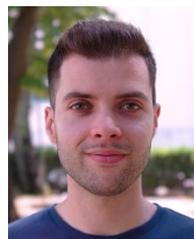

**ANDRÉ F. R. GUARDA** (Member, IEEE) received his B.Sc. and M.Sc. degrees in Electrotechnical Engineering from Instituto Politécnico de Leiria, Portugal, in 2013 and 2016, respectively, and the Ph.D. degree in Electrical and Computer Engineering from Instituto Superior Técnico, Universidade de Lisboa, Portugal, in 2021. He has been a researcher at Instituto de Telecomunicações since 2011, where he currently holds a Post-Doctoral position. His main research interests include multimedia signal processing and coding, with particular focus on point cloud coding with deep learning. He has authored several publications in top conferences and journals in this field, and is actively contributing to the standardization efforts of JPEG and MPEG on learning-based point cloud coding.

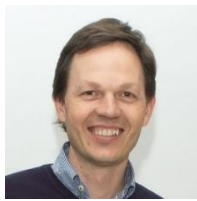

**NUNO M. M. RODRIGUES** (Senior Member) graduated in electrical engineering in 1997, received the M.Sc. degree from the Universidade de Coimbra, Portugal, in 2000, and the Ph.D. degree from the Universidade de Coimbra, Portugal, in 2009, in collaboration with the Universidade Federal do Rio de Janeiro, Brazil. He is a Professor in the Department of Electrical Engineering, in the School of Technology and Management of the Polytechnic University of Leiria, Portugal and a Senior Researcher in Instituto de Telecomunicações, Portugal. He has coordinated and participated as a researcher in various national and international funded projects. He has supervised three concluded PhD theses and several MSc theses. He is co-author of a book and more than 100 publications, including book chapters and papers in national and international journals and conferences. His research interests include several topics related with image and video coding and processing, for different signal modalities and applications. His current research is focused on deep learning-based techniques for point cloud coding and processing.

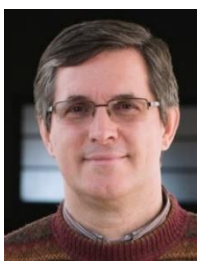

**FERNANDO PEREIRA** (Fellow, IEEE) graduated in electrical and computer engineering in 1985 and received the M.Sc. and Ph.D. degrees in 1988 and 1991, respectively, from Instituto Superior Técnico, Technical University of Lisbon. He is currently with the Department of Electrical and Computer Engineering, Instituto Superior Técnico and Instituto de Telecomunicações, Lisbon, Portugal. He is also the JPEG Requirements Subgroup Chair. Recently, he has been one of the key designers of the JPEG Pleno and JPEG AI standardization projects. He has contributed more than 300 papers in international journals, conferences, and workshops, and made several tens of invited talks and tutorials at conferences and workshops. His research interests include visual analysis, representation, coding, description and adaptation, and advanced multimedia services. He was an IEEE Distinguished Lecturer, in 2005, and elected as a fellow of IEEE in 2008 for "Contributions to object-based digital video representation technologies and standards." Since 2013, he has been a EURASIP Fellow for "Contributions to digital video representation technologies and standards." Since 2015, he has been a fellow of IET. He is or has been a member of the Editorial Board of the Signal Processing Magazine, an Associate Editor of IEEE TRANSACTIONS ON CIRCUITS AND SYSTEMS FOR VIDEO TECHNOLOGY, IEEE TRANSACTIONS ON IMAGE PROCESSING, IEEE TRANSACTIONS ON MULTIMEDIA, and IEEE Signal Processing Magazine. He has been elected to serve on the Signal Processing Society Board of Governors in the Capacity of Member-at-Large (2012) and (2014–2016). He has been the Vice President of the IEEE Signal Processing Society, from 2018 to 2020. He has also been elected to serve on the European Signal Processing Society Board of Directors (2015–2018). He is an Area Editor of the Signal Processing: Image Communication journal and an Associate Editor of the EURASIP Journal on Image and Video Processing. From 2013 to 2015, he was the Editor-in-Chief of the IEEE JOURNAL OF SELECTED TOPICS IN SIGNAL PROCESSING.